# Monolingual and Parallel Corpora for Kangri Low Resource Language

Shweta Chauhan, Shefali Saxena, Philemon Daniel

National Institute of Technology, Hamirpur, Himachal Pradesh, INDIA

{shweta, shefali, phildani7)@nith.ac.in

## Abstract

In this paper we present the dataset of Himachali low resource endangered language, Kangri (ISO 639-3xnr) listed in the United Nations Educational, Scientific and Cultural Organization (UNESCO). The compilation of kangri corpus has been a challenging task due to the non-availability of the digitalized resources. The corpus contains 1,81,552 Monolingual and 27,362 Hindi-Kangri Parallel corpora. We shared pre-trained kangri word embeddings. We also reported the Bilingual Evaluation Understudy (BLEU) score and Metric for Evaluation of Translation with Explicit ORdering (METEOR) score of Statistical Machine Translation (SMT) and Neural Machine Translation (NMT) results for the corpus. The corpus is freely available for non-commercial usages and research. To the best of our knowledge, this is the first Himachali low resource endangered language corpus. The resources are available at (https://github.com/chauhanshweta/Kangri_corpus)

## 1 Introduction

Compilation of data for high resource languages is not a challenge anymore due to their widespread digital presence and usage in various Natural Language Processing (NLP) applications. Standard Corpus have been designed for these high resource languages, but under-resourced languages are still lagging in making their digital presence.

Low resource languages are those which have very less or insignificant datasets for applying supervised learning algorithms. There are 2,464 endangered language around the world listed in the UNESCO atlas of world's languages in danger (UNESCO, 2017). India tops the list with 197 endangered languages, followed by the U.S. (191) and Brazil (190). Out of 197 languages, 81 are vulnerable followed by 63 definitely endangered, 6 severely endangered, 42 critically endangered. In definitely endangered language, children no longer learn the language as their mother tongue.

India is a very diverse country in the context of religion, culture, language, art etc. Himachal Pradesh, an Indian state, has 7 definitely endangered languages, kangri being the one. Kangri is an Indo-Aryan language variety spoken in the Kangra, Hamirpur and Una districts of Himachal Pradesh. According to the 2011 Census, Kangri is spoken by around 1.7 million people in India, of small fraction of fluent speakers (Wikipedia, 2020).

| Corpus Id | Sources | Number of segments |
|---|---|---|
| **Kr_1** | Books | 1,06117 |
| **Kr_2** | Hindi-Kangri dictionary | 9,082 |
| **Kr_3** | Poems | |
| | • Kavitayein | 50,991 |
| | • Lok-Geet | 12,486 |
| | • Gazals | 2,443 |
| Total | | 1,81,552 |

Table 1: Details of the Monolingual kangri corpus

Since kangri language is not used as an official language in offices, tourism, education etc. It is deteriorating every year. Hence, there is serious or immediate requirement for enhancing everyday usages of the language. Towards this purpose, digitalization of this language and developing machine translation models, digital dictionaries etc are of atmost important. The difficulty increases as kangri is morphologically rich languages, and they do not have well defined linguistic rules. Compiling a corpus is hard because dialect varies slightly from region to region.

In this paper we have presented kangri monolingual and Hindi-kangri parallel dataset. Kangri is written in Devanagari script, which is commonly used for Indo-Aryan languages. We have manually compiled and organized Kangri dataset. Hence, monolingual dataset consists of 1,81,552 sentences and parallel dataset contain 27,362 sentences. Pre-trained kangri word embeddings by using fastText is also shared. We have included the results of BLEU and METEOR score for SMT and NMT translations.

The rest of paper is organized as: Section 2 describe the dataset collection. Kangri word embedding are presented in Section 3. Section 4 include machine translation and section 5 concludes the paper along with future work.

## 2 Corpus Collection

As there is almost no online Kangri resources available, we have manually compiled and created the dataset as explained in the section below.

### 2.1 Compilation of Monolingual Corpus

For monolingual corpus compilation, primary sources are the kangri books that were collected from kangri authors. We have digitized the books and divided them into number of sentences to generate this dataset. For digitization of kangri books we have used Optical Character Recognition (OCR) base approach, we found a few errors such as unnecessary characters and symbols, missing words, etc. Although this task is time-consuming and tedious, we have manually corrected the sentences.

In Table 1 the monolingual corpus is divided into three categories as Kr_1, Kr_2 and Kr _3. The Kr_1 contains Kangri stories, Kr_2 presents the Hindi-Kangri dictionaries, Kr_3 contains Poems.

| Corpus Id | Dataset | Sentences |
|---|---|---|
| **Kr_4** | Hindi-kangri (Parallel Dataset) | 27,362 |
| Total | Total | 27,362 |

Table 2: Details of parallel Hindi-Kangri corpus

| Dataset | | Sentence | Token | Vocabulary | Train | Test |
|---|---|---|---|---|---|---|
| Monolingual | Kangri | 1,81,552 | 2377100 | 223153 | 1,80,552 | 1,000 |
| Parallel | Hindi | 27,362 | 281076 | 25290 | 26,862 | 500 |
| | Kangri | 27,362 | 271752 | 43504 | 26,862 | 500 |

Table 3: Statistics of the datasets

**Kangri stories:** The collected books contain various short/long stories and novels. Apart from books we have also compiled the monolingual data by including conversations from various WhatsApp and Facebook groups.

**Hindi-Kangri dictionary:** This section includes Hindi-kangri dictionary words.

**Poems:** The Kavitaiyein, Lok-Geet and kangri Gazals written by various kangri authors.

### 2.2 Creation of Parallel Hindi-Kangri Dataset

The Parallel Hindi-Kangri dataset are shown in Table 2. The parallel corpus has been created by distributing different everyday topics to kangri writers. Both Hindi and kangri were written from scratch. The main categories are as follows:

- Hospital, Defense, Media, School, Music, Sports, Dance, Food, Parties, Law, Market, Marriage, Culture, History, Education, Technology, Religion, Stories, etc.

### 2.3 Corpus Statistics

The statistics of monolingual dataset are shown in Table 3. There are 1,92,000 sentences and 2,23,153 tokens. Out of the 1,78,552 sentences were randomly picked for training and 1000 sentences as test dataset for monolingual kangri dataset. Similarly, for parallel datasets there is 25,362 are training dataset whereas 500 is test dataset.

### 3 Kangri Word Embedding

Word embedding is the vector representation of each word which is capable of capturing word context, syntactic and semantic similarity, word analogies and its relationship with other words.

FastText (Joulin et al.,2016) approach based on SkipGram and CBOW model which represent each word as bag of n-gram characters. We have tokenized the corpus using FastText because of its efficiently better word relationship.

### 3.1 Kangri Word Embedding Nearest Neighbors

Here, we have shown the prediction of nearest top five values for kangri word with their similarity score. We have selected some common source words as मने, भाऊ, मिंजो, कुथु and their best five nearest kangri words are shown in Table 4.

### 3.2 Kangri Word Embedding Visualization

Principal Component Analysis (PCA) is a

| Source Word | | मने | | भाऊ | | मिंजो | | कुथु | |
|---|---|---|---|---|---|---|---|---|---|
| Sr. No | Value | Word | Value | Word | Value | Word | Value | Word | |
| 1. | 1.000 | मने | 1.000 | भाऊ | 1.000 | मिंजो | 1.000 | कुथु | |
| 2. | 0.7314 | मन | 0.8296 | भाऊय | 0.5669 | मींजो | 0.7478 | कुतु | |
| 3. | 0.7029 | सुपने | 0.8043 | भाऊआ | 0.5550 | मिंहजो | 0.7006 | कताहां | |
| 4. | 0.6885 | मनोए | 0.7865 | भाऊ | 0.5390 | मिजो | 0.6063 | क्त्ता | |
| 5. | 0.6643 | मनें | 0.5820 | भाग | 0.5387 | मिज्जों | 0.5936 | कुथह | |

Table 4: Nearest Neighbour Similarity Score Translation of kangri

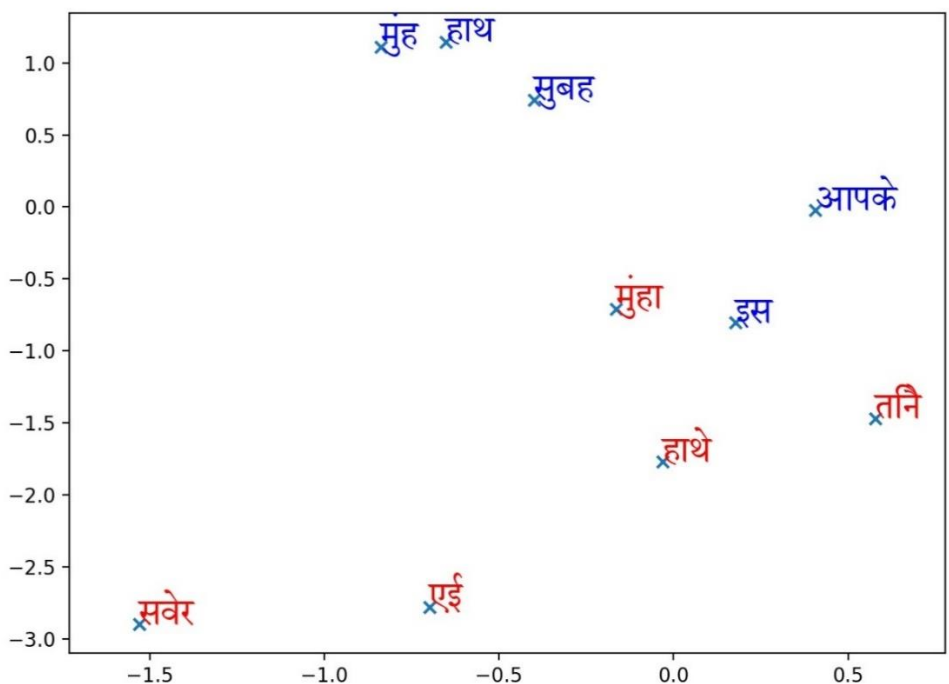

Fig 1. A shared embedding space between Hindi and Kangri.

dimensionality reduction method that is often used to reduce the dimensionality of large data sets, by transforming a large set of variables into a smaller one that still contains most of the information in the large set. It is often used to make data easy to explore and visualize. We used FastText with a skip-gram model (Joulin et al.,2016) for the generation of word vectors. The training parameters for word embedding is set as 10 epochs, learning rate is 0.05 and word embedding dimension as 300.

Fig 1 shows the word embedding visualization in two-dimensional space, we have plotted some of Kangri words in the mapped area using the nearest neighbor concept. Some random words are taken from different domains and their dimensionality is reduced using PCA. As in Fig. 1 some Hindi-Kangri words like 'मुंह' and 'मुंहा' are very close to each other, therefore, can be easily translated from one language to another language.

| Source Word | तुम | | जमींदार | | लड़की | | क्यों | |
|---|---|---|---|---|---|---|---|---|
| Sr. No | Value | Word | Value | Word | Value | Word | Value | Word |
| 1. | 0.5922 | तुहां | 0.5546 | सहूंकार | 0.6888 | कुड़ी | 0.6723 | कजो |
| 2. | 0.5875 | तिना | 0.5448 | गणसु | 0.6669 | कुड़िए | 0.6478 | क्ऱज्जो |
| 3. | 0.5812 | तिहनां | 0.5402 | रुपये | 0.6550 | मुन्नी | 0.6236 | कछ |
| 4. | 0.5797 | खेतारां | 0.5394 | गुडशय | 0.6390 | काकी | 0.6063 | कुछ |
| 5. | 0.5577 | जमनां | 0.5353 | गणूँसुए | 0.5987 | मुनिया | 0.5936 | खज्जी |

Table 5: Nearest Neighbor of Hindi to kangri Word Translation.

| Sr. Number | Translation Models | Kangri-Hindi | | | Hindi-kangri | |
|---|---|---|---|---|---|---|
| | | BLEU | METEOR | METEOR-Hindi | BLEU | METEOR |
| 1. | NMT | 3.25 | 1.50 | 3.92 | 4.23 | 1.98 |
| 2. | SMT | 4.98 | 1.92 | - | 5.75 | 2.01 |

Table 6: Evaluation Score for Kangri and Hindi

### 3.3 Word to Word Translation

Cross-lingual word embedding (CLWE) tries to align the word embeddings trained on source and target corpora in common n-dimensional embedding space such that they are associated with each other (Artetxe et al., 2018). CLWE represents continuous words in vector or real numbers in a vector space that is shared across multiple languages. This helps in measuring distance between word embeddings across multiple languages, for finding possible word translation. Firstly, the learning both source and target embedding from their monolingual corpora is done independently. Secondly, source embedding space is linearly mapped with the target embedding space by fully unsupervised robust self-learning method (Artetxe et al., 2018).In this section, we are predicting the best five values for Hindi source words to the Kangri target space. We have randomly selected source words तुम, जमींदार, लड़की, क्यों and the top five nearest kangri words are shown in Table 5.

## 4 Machine translation

### 4.1 NMT Model

We used the shared encoder with backtranslation (Artetxe et al., 2017)that employs one shared encoder with two separate decoders for both the languages. We have used 2 layered encoder decoder with (Vashwani et al,2017) ,600 hidden units each, GRU cells, batch size of 50 and maximum sentence length of 50. Adam (Kingma and Ba, 2014) optimizer is used with learning rate of 0.0002. We have trained the system with fixed 300000 iterations. Total training time on the Titan XP GPU is for 2 days.

### 4.2 SMT Model

We used Phrase Base SMT (PBSMT) system provided by the Moses toolkit (Koehn et al.,2007) for training the PBSMT statistical machine translation systems using monolingual corpora (Artetxe et al.,2018). The word segmented source language was aligned with the word segmented target language using FastAlign (Dyer et al.,2013). The alignment was symmetrize by grow-diag-final and heuristic (Koehn et al.,2003). We use KenLM (Heafield, Kenneth 2011) for training the 5-gram language model with modified Kneser-Ney discounting (Chen Stanley F and Goodman Joshua 1996)). Minimum error rate training (MERT) (Och Franz J et al.,2003) was used to tune the Unsupervised SMT model. We used default settings of Moses (Koehn et al., 2007) for all the experiments.

### 4.3. Evaluation Score

We evaluated machine translated output using BLEU (Papineni et al.,2002), METEOR (Banerjee and Lavie, 2005) and METEOR-Hindi (Gupta et al., 2010) evaluation metrics. Table 6 shows the results of our experiments.

### 5. Conclusion and Future Work

In this paper, we have built a primary corpus for the kangri language. We believe that this will enrich the usages of kangri language and translation tasks with kangri as source or target.

We presented the kangri monolingual and parallel corpus version 1.0 and provided SMT and NMT results on this corpus. The corpus is available under a creative commons license. In future, we plan to enhance the corpus and provide other low resources definitely endangered Himachali languages datasets. We hope the availability of this corpus will accelerate NLP research for Indian languages by enabling the community to build further resources and solutions for various NLP tasks and opening interesting NLP questions.

### Acknowledgements

We thank Dr. Karam Singh, Director of Department of Language Art and Culture, Shimla, Himachal Pradesh for their efforts in arranging workshops to collect datasets. We also thank to all the authors of various books. We also thank all the language translators writers who manually compiles the dataset: Aman Kumar Vishva, Bharti kudailiya, Bhupinder Singh Bhupi, Deepak kulavi, Vijay Puri , Hari Krishan Murari ,Navin Haldwani , Suresh Lata Awasthi ,Shakti Singh Rana, Shelly Kiran, Vandana Rana, Vinod Bhavuk , Virender Sharma, Gopal Sharma, Naveen, Rajiv Trigti, Manoj Kumar.